\documentclass[conference]{IEEEtran}
\IEEEoverridecommandlockouts
\usepackage{cite}
\usepackage{amsmath,amssymb,amsfonts}
\usepackage{algorithmic}
\usepackage{graphicx}
\usepackage{textcomp}
\usepackage{xcolor}
\usepackage{tikz}
\usepackage{color}
\usepackage{pgfpages}
\usepackage{pgfplots}
\usepackage{algorithm}
\usepackage{algorithmic}
\def\BibTeX{{\rm B\kern-.05em{\sc i\kern-.025em b}\kern-.08em
    T\kern-.1667em\lower.7ex\hbox{E}\kern-.125emX}}
\begin{document}
\author{Ghouthi B. HACENE $^{1,2}$, Vincent GRIPON $^{1,2}$, Nicolas FARRUGIA$^{2}$, Matthieu ARZEL $^{2}$ and Michel JEZEQUEL$^{2}$\\
        $^1$ Universit\'e de Montr\'eal, MILA 
        \hspace{5pt} $^2$ IMT Atlantique, Lab-STICC\\

        }
\title{Efficient Hardware Implementation of\\Incremental Learning and Inference on Chip}

\maketitle

\begin{abstract}
  In this paper, we tackle the problem of incrementally learning a classifier, one example at a time, directly on chip. To this end we propose an efficient hardware implementation of a recently introduced incremental learning procedure that achieves state-of-the-art performance by combining transfer learning with majority votes and quantization techniques. The proposed design is able to accommodate for both new examples and new classes directly on the chip. We detail the hardware implementation of the method (implemented on FPGA target) and show it requires limited resources while providing a significant acceleration compared to using a CPU.
\end{abstract}


\section{Introduction}
\label{sec:introduction}

Deep Neural Networks (DNNs) based methods offer state of the art performance in many domains such as computer vision and speech recognition~\cite{lecun2015deep}. They rely on large quantities of available data and hundreds of millions of trainable parameters, which require significant memory capacities and computational power. As a consequence, their implementation on embedded systems is a challenge. Several approaches have been introduced aiming at reducing DNNs size and complexity, such as using product quantization (PQ) to factorize DNNs weights~\cite{wu2016quantized},\cite{gong2014compressing}, binarizing both DNNs weights and activations~\cite{courbariaux2015binaryconnect,soulie2016compression,choi2016towards}, pruning DNN connections~\cite{yamamoto2018pcas, huang2018learning}, or replacing the spatial convolution by a shift operation followed by $1 \times 1$ convolution~\cite{wu2017shift,jeon2018constructing}. These methods ease the implementation of big and complex DNNs on embedded systems~\cite{qiu2016going,suda2016throughput,ardakani2018convolutional} as far as the inference part is concerned. However, because it requires storing the whole dataset and causes increased complexity, the training of DNNs is in most cases still performed offline.

In the literature multiple works have tackled the question of Learning On Chip (LOC)~\cite{bo2000chip,paul2006back,ortega2016efficient,lacey2016deep}. In the context of deep learning, a key problem is the use of the stochastic gradient descent algorithm, that requires accessing the training dataset multiple times, and thus storing all of it in memory. Methods that do not require storing the whole dataset in memory are referred to as ``incremental learning'' in the literature. An incremental learning~\cite{iSVM,poggio2001incremental,learn++,tweet,rebuffi-cvpr2017,hacenebudget,mensink2013distance,hacene2017incremental} solution is defined as learning sequentially one (or few) example(s) at a time without the need to access previously processed data. Unfortunately the incremental part of the learning process comes with a loss in accuracy that might be too damageable for some applications.

Recently authors have proposed a solution named Transfer Incremental Learning with Data Augmentation (TILDA) to considerably increase the accuracy of incremental leaning techniques by combining transfer learning with quantization techniques and majority votes~\cite{hacene2018transfer}. The term ``transfer learning'' refers to the the use of pre-trained DNNs to obtain new representations of input signals~\cite{girshick2014rich,pan2010survey}. 
In this work, we propose a hardware implementation of TILDA~\cite{hacene2018transfer}. We show that the proposed hardware solution requires limited resources while providing substantial gains in processing time compared to a CPU. We also show in Table~\ref{table:acc} that the method is able to compete with state-of-the-art non-incremental transfer learning alternatives.

\begin{figure*}
\centering
\begin{tikzpicture}[scale=0.9]
\def \xsmal{0.2}
\node[text width=3cm] at (-5.6,-1.2) 
    {};
\draw[->] (-3.6,-1.5) -- (-2.6,-1.5);
\draw (-2.9,-1.3)--(-3.3,-1.7);
\node[text width=3cm] at (\xsmal+-2.2,-0.6) 
{\small \textbf{Feature \\Vector}};
\node[text width=3cm] at (\xsmal+-1.8,-1.2) 
    {\small $\textbf{X}^m$};
\node[text width=3cm] at (\xsmal+-2.1+.1,-1.8-.1) 
    {\small \textit{$nT$}};

\draw[->] (-3.6,-5.7) -- (-.4,-5.7);
\draw (-2.9,-5.5)--(-3.3,-5.9);
\node[text width=3cm] at (\xsmal+-2,-5) 
{\small \textbf{Input \\Class}};
\node[text width=3cm] at (\xsmal+-1.7,-6) 
    {\small \textit{$n$}};

\fill[green] (-2.5,-4.5) -- (-2.5,1.5) -- (-.5,1.5)-- (-.5,-4.5)-- cycle;
\node[text width=3cm] at (\xsmal+-.4,-1) 
    {\small Input \\Register};

\draw[->] (-.4,.9)--(.9,.9);
\draw (.3,1.1)--(.1,.7);

\draw[->] (-.4,-1.1)--(.9,-1.1);
\draw (.1,-1.3)--(.3,-.9);
\draw[->] (-.4,-3.1)--(.9,-3.1);
\draw (.1,-3.3)--(.3,-2.9);

\node[text width=3cm] at (\xsmal+1.1,0.4) 
    {\small \textit{$nT/P$}};
\node[text width=3cm] at (\xsmal+1.1,-1.6) 
    {\small \textit{$nT/P$}};
\node[text width=3cm] at (\xsmal+1.1,-3.6) 
    {\small \textit{$nT/P$}};

\draw (-2.5,-6.7)--(-1.4,-6.7)--(-1.4,-6)--(-2.5,-6)--cycle;
\node[text width=3cm] at (\xsmal+-1,-6.2) 
    {\small counter};
\node[text width=3cm] at (\xsmal+-.7,-6.5) 
    {\small L-P};
    
\draw[->] (-3.6,-6.3)--(-2.6,-6.3);
\node[text width=3cm] at (\xsmal-2.9,-6.3) 
    {\small \textbf{L-P}};
    

\node[text width=3cm] at (\xsmal+.5,-6.7) 
    {\small \textit{$n$}};
    
\draw (-.3,-5.5)--(-.3,-6.5)--(.8,-6.5)--(.8,-5.5)--cycle;
\node[text width=3cm] at (\xsmal+1.35,-6) 
    {\small ADD};
    
\node[text width=3cm] at (\xsmal+2.55,-5.5) 
    {\small \textbf{Address}};
    
\draw[->] (-1.3,-6.35)--(-.4,-6.35);
\draw(-.7,-6.15)--(-1,-6.55);
\node[text width=3cm] at (\xsmal+.5,-6.7) 
    {\small \textit{$n$}};

\draw(.9,-6)--(3,-6);
\draw(3,-6)--(3,-3.3);
\draw[dashed] (3,-3.3)--(3,-2.7);
\draw(3,-2.7)--(3,-1.3);
\draw[dashed] (3,-1.3)--(3,-.7);
\draw(3,-.7)--(3,0);

\draw(1.5,-5.8)--(1.2,-6.2);
\node[text width=3cm] at (\xsmal+2.7,-6.3) 
    {\small \textit{$n$}};

\draw (1.1,.3)--(2.7,.3)--(2.7,1.6)--(1.1,1.6)--cycle;
\node[text width=3cm] at (\xsmal+2.6,1.1) 
    {\small Processing};
    \node[text width=3cm] at (\xsmal+3.05,0.7) 
    {\small block};
\draw[->](3,0)--(1.9,0)--(1.9,.2);

\draw (1.1,-1.7)--(2.7,-1.7)--(2.7,-.4)--(1.1,-.4)--cycle;
\node[text width=3cm] at (\xsmal+2.6,-.9) 
    {\small Processing};
    \node[text width=3cm] at (\xsmal+3.05,-1.3) 
    {\small block}; 
\draw[->](3,-2)--(1.9,-2)--(1.9,-1.8);

\draw (1.1,-3.7)--(2.7,-3.7)--(2.7,-2.4)--(1.1,-2.4)--cycle;
\node[text width=3cm] at (\xsmal+2.6,-2.9) 
    {\small Processing};
\node[text width=3cm] at (\xsmal+3.05,-3.3) 
    {\small block};
\draw[->](3,-4)--(1.9,-4)--(1.9,-3.8);

\draw[->] (2.8,.9)--(3.7,.9);
\draw (3.4,1.1)--(3.1,.7);

\draw[->] (2.8,-1.1)--(3.7,-1.1);
\draw (3.1,-1.3)--(3.4,-.9);
\draw[->] (2.8,-3.1)--(3.7,-3.1);
\draw (3.1,-3.3)--(3.4,-2.9);

\node[text width=3cm] at (\xsmal+4.6,0.4) 
    {\small \textit{$C$}};
\node[text width=3cm] at (\xsmal+4.6,-1.6) 
    {\small \textit{$C$}};
\node[text width=3cm] at (\xsmal+4.6,-3.6) 
    {\small \textit{$C$}};



\draw(3.8,1.4)--(5.6,1.4)--(5.6,-4.2)--(3.8,-4.2)--cycle;
\node[text width=3cm] at (\xsmal+5.35+.2,-1) 
    {\small Parallel\\Majority\\Vote};
\draw[->](5.7,-1.5)--(6.7,-1.5);
\draw(6.1,-1.7)--(6.4,-1.3);
\node[text width=3cm] at (\xsmal+7.2-.1,-0.8) 
    {\small \textbf{Output\\Class}};
\node[text width=3cm] at (\xsmal+7.5,-2) 
    {\small $C$};

\draw(6.8,1.4)--(8.6,1.4)--(8.6,-4.2)--(6.8,-4.2)--cycle;
\node[text width=3cm] at (\xsmal+8.35+.1,-1) 
    {\small Sequential\\Majority\\Vote};
\draw[->](8.7,-1.5)--(9.7,-1.5);
\draw(9.1,-1.7)--(9.4,-1.3);
\node[text width=3cm] at (\xsmal+10.2,-0.8) 
    {\textbf{Output\\Class}};
\node[text width=3cm] at (\xsmal+10.5,-2) 
    {\textit{$C$}};


\end{tikzpicture}
\caption{Hardware architecture for incremental learning.}
\label{fig:HDfig}
\vspace{-.5cm}
\end{figure*}
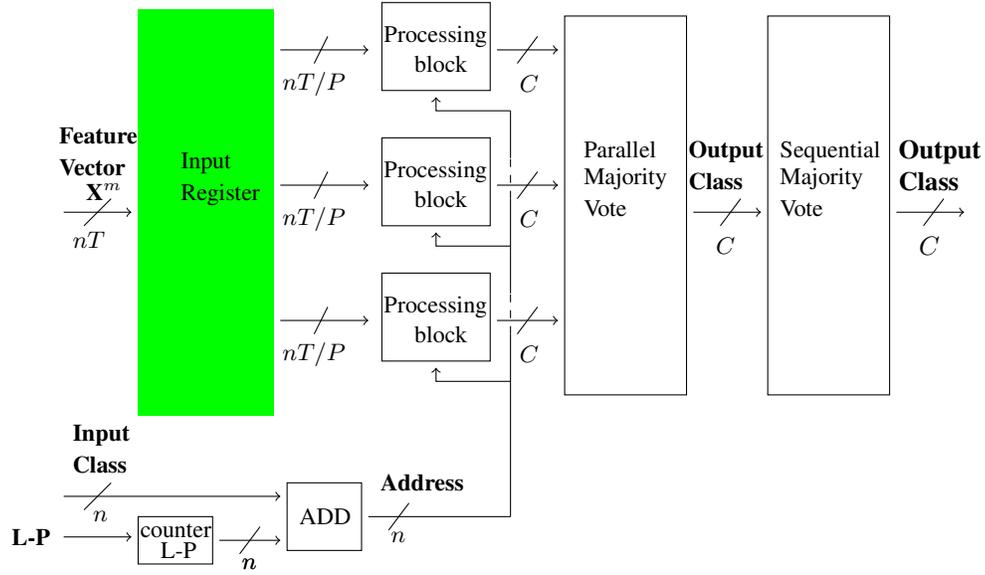

\section{Proposed Method}
\label{sec3}
Let us first introduce notations (extracted from~\cite{hacene2018transfer}). TILDA is made of four main ingredients: 1) a pre-trained and fixed DNN to perform feature extraction of signals, 2) a vector splitting procedure to project features into low dimensional subspaces, 3) an assembly of NCM-inspired classifiers applied independently in each subspace and 4) a data augmentation scheme to increase accuracy of the classifying process.

The first step consists in performing transfer using the internal layers of a pre-trained DNN acting as a generic feature extractor to compute the feature vector $\textbf{x}^m$, corresponding to the input signal $\textbf{s}^m$. Since there is already a lot of literature on the subject of hardware implementation of the inference of DNNs~\cite{qiu2016going,suda2016throughput,ardakani2018convolutional}, we disregard this first step in the implementation described in this paper and directly consider processing the vector $\textbf{x}^m$. 

Then, each feature vector $\textbf{x}^m$ is split into $P$ subvectors of equal size denoted $\left(\textbf{x}_p^m\right)_{1\leq p \leq P}$.
During training, for each class $c$ and each subspace $p$, we produce $k$ anchor vectors $Y_{c,p}=[\textbf{y}_{c,p,1},..., \textbf{y}_{c,p,k}]$  initialised  with $0$s, and their associated counters $N_{c,p}=[n_{c,p,1},\dots, n_{c,p,k}]$ also initialised by $0$s.

Then, each time an input training vector is processed, an anchor vector is identified to be updated. The update simply consists of computing a new anchor vector obtained as a barycenter of the old one with weight given by its counter and the input subvector with weight 1, then incrementing  the counter. This procedure is detailed in Algorithm~\ref{Algo:one}.

\begin{algorithm}
\caption{Incremental Learning of Anchor Subvectors}
 \textbf{Input}: streaming feature vector $\textbf{x}_c^{m}$\\
\begin{algorithmic}

\FOR{$p:=1$ to $P$}
\FOR{$i:=1$ to $k$}
\STATE $d_i=\| \textbf{x}_{p} - \textbf{y}_{c,p,i}\|_2$
\STATE $R_i=d_i n_{c,p,i}$

\ENDFOR
\STATE $\tilde{k}=\displaystyle{\arg\min_{i}R_i}$
\STATE $\textbf{y}_{c,p,\tilde{k}} \leftarrow \textbf{y}_{c,p,\tilde{k}} n_{c,p,\tilde{k}} + \textbf{x}_{c,p}^{m}$
\STATE $n_{c,p,\tilde{k}} \leftarrow n_{c,p,\tilde{k}} + 1$
\STATE $\textbf{y}_{c,p,\tilde{k}} \leftarrow \textbf{y}_{c,p,\tilde{k}}/n_{c,p,\tilde{k}}$
\ENDFOR

\end{algorithmic}
\label{Algo:one}
\end{algorithm}

During the prediction phase, for each subspace, anchor vectors are divided by their corresponding counters and used for nearest-neighbour search. More precisely, each subpart $\mathbf{x}_p$ obtained from input signal $\mathbf{s}$ is weakly classified using the nearest-neighbour anchor vector in part $p$. Finally, a majority vote is performed to aggregate these weak classifications.

To further improve the method performance, data augmentation is used at both prediction and training phases. The training data-augmentation is used to artificially enrich the training dataset, whereas the prediction data-augmentation is used to obtain multiple decisions that are aggregated using a second majority vote.

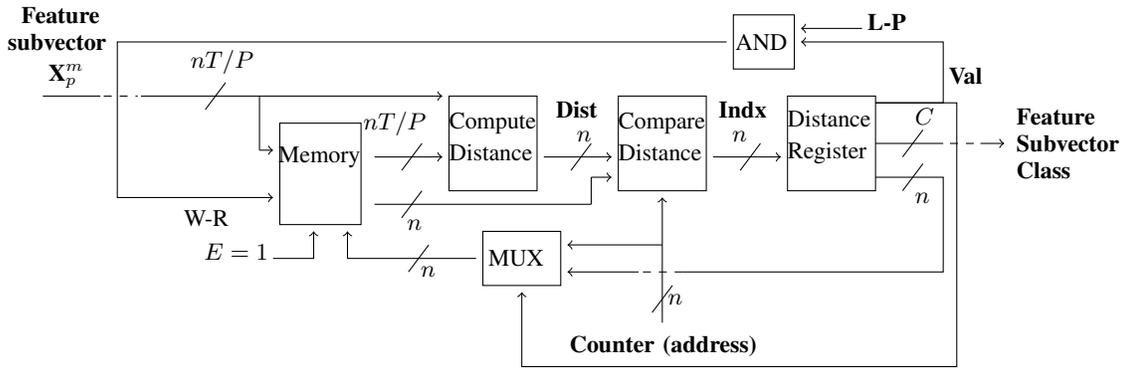
\begin{figure*}
\centering
\begin{tikzpicture}[scale=0.9]
\def \xsmal{0.15}
\draw[->](8.95,-1.95)--(8.95,-.1);
\draw (9.1,-1.4)--(8.8,-1.8);
\node[text width=3cm] at (-1,-1.6) 
    {};
\node[text width=3cm] at (10.7,-1.6) 
    {\small \textit{$n$}};
    
\draw[->](8.95,-.8)--(7.55,-.8);
\draw (7.45,-.6)--(6.3,-.6)--(6.3,-1.4)--(7.45,-1.4)--cycle;
\node[text width=3cm] at (7.9+\xsmal,-1) 
    {\small MUX};

\draw[->](12.1,1.3)--(13.3,1.3)--(13.3,-2.6)--(6.875,-2.6)--(6.875,-1.5);
\draw[->](6.2,-1)--(4.3,-1)--(4.3,-.6);
\draw(5.2,-1.2)--(5.5,-.8);
\node[text width=3cm] at (\xsmal+6.9,-1.2) 
    {\small \textit{$n$}};
    
\draw[->](3.2,-1)--(3.8,-1)--(3.8,-.6);
\node[text width=3cm] at (\xsmal+3.7,-.9) 
    {\small $E=1$};


\draw (3.3,-0.5)--(4.5,-0.5)--(4.5,1)--(3.3,1)--cycle;
\node[text width=3cm] at (\xsmal+4.8,0.5) 
    {\small Memory};
    
\draw[->] (1.25,1.4)--(5.7,1.4);
\draw[dashed] (.7,1.4)--(1.2,1.4);
\draw (-.2,1.4)--(.65,1.4);
\draw (2.5,1.6)--(2.2,1.2);
\node[text width=3cm] at (\xsmal+3.5,1.9) 
    {\small \textit{$nT/P$}};
\node[text width=3cm] at (\xsmal+1,2.6) 
    {\small \textbf{Feature}};
\node[text width=3cm] at (\xsmal+.8,2.2) 
    {\small \textbf{subvector}};
\node[text width=3cm] at (\xsmal+1.4,1.7) 
    {\small $\textbf{X}^m_p$};
\draw (3,1.4) -- (3,0.6);
\draw[->](3,0.6)--(3.2,.6);
\draw[->](4.7,0.5)--(5.7,0.5);
\draw(5.1,0.3)--(5.4,0.7);
\node[text width=3cm] at (\xsmal+6.05,1) 
    {\small \textit{$nT/P$}};

\draw(4.7,-0.2)--(7.9,-0.2);
\draw(7.9,-0.2)--(7.9,0.2);
\draw[->](7.9,0.2)--(8.2,0.2);
\draw(5.4,0)--(5.1,-0.4);
\node[text width=3cm] at (\xsmal+6.7,-.5) 
    {\small \textit{$n$}};

    
\draw (5.8,-0)--(7.1,-0)--(7.1,1.4)--(5.8,1.4)--cycle;
\node[text width=3cm] at (\xsmal+7.32,0.8) 
    {\small Compute\\Distance};
\draw[->](7.2,0.5)--(8.2,0.5);
\draw(7.6,0.3)--(7.9,0.7);
\node[text width=3cm] at (\xsmal+9.2,.8) 
    {\small \textit{$n$}};
\node[text width=3cm] at (\xsmal+8.9,1.2) 
    {\small \textbf{Dist}};
\draw (8.3,-0)--(9.6,-0)--(9.6,1.4)--(8.3,1.4)-- cycle;
\node[text width=3cm] at (\xsmal+9.82,0.8) 
    {\small Compare\\Distance};
\draw[->](9.7,0.5)--(10.7,0.5);
\draw(10.1,0.3)--(10.4,0.7);
\node[text width=3cm] at (\xsmal+11.5,0.8) 
    {\small \textit{$n$}};
\node[text width=3cm] at (\xsmal+11.3,1.2) 
    {\small \textbf{Indx}};
\node[text width=3cm] at (\xsmal+9.1,-2.3) 
    {\small \textbf{Counter (address)}};
\draw (10.8,-0)--(12.1,-0)--(12.1,1.4)--(10.8,1.4)-- cycle;
\node[text width=3cm] at (\xsmal+12.32,0.8) 
    {\small Distance\\Register};
\draw(12.1,.7)--(13.1,.7);
\draw[dashed](13.2,.7)--(13.6,.7);
\draw[->](13.65,.7)--(14,.7);
\draw(12.5,.5)--(12.8,.9);

\node[text width=3cm] at (\xsmal+14.2,1.1) 
    {\small \textit{$C$}};
\node[text width=3cm] at (\xsmal+15.7,1.1) 
    {\small \textbf{Feature}};
\node[text width=3cm] at (\xsmal+15.7,.7) 
    {\small \textbf{Subvector}};
\node[text width=3cm] at (\xsmal+15.7,.3) 
    {\small \textbf{Class}};
\draw(12.1,.2)--(13.1,0.2)--(13.1,-1.2)--(9.25,-1.2);
\draw[dashed](9.25,-1.2)--(8.6,-1.2);
\draw[->](8.6,-1.2)--(7.55,-1.2);
\draw(12.5,.0)--(12.8,.4);
\node[text width=3cm] at (\xsmal+14.2,-.1) 
    {\small \textit{$n$}};
\draw[->](12.1,1.3)--(13.1,1.3)--(13.1,2.2)--(11,2.2);
\draw[->](11.9,2.4)--(11,2.4);
\node[text width=3cm] at (\xsmal+13.5,2.5) 
    {\small \textbf{L-P}};
\node[text width=3cm] at (\xsmal+14.7,1.7) 
    {\small \textbf{Val}};
\draw(10.9,1.8)--(10.9,2.6)--(10,2.6)--(10,1.8)--cycle;
\node[text width=3cm] at (\xsmal+11.55,2.2) 
    {\small AND};

\draw[->](9.9,2.2)--(.9,2.2)--(.9,-0.1)--(3.2,-0.1);

\node[text width=3cm] at (\xsmal+3.4,-.4) 
    {\small W-R};

\end{tikzpicture}
\caption{Hardware architecture of Processing block.}
\label{fig: Subspace}
\end{figure*}

\section{Hardware Implementation}
\label{sec4}
In this paper, we assume that a generic feature extraction is performed by an external CPU, and provides feature vectors $\mathbf{x}^m$ to the FPGA. Consequently, we introduce a hardware implementation to compute the incremental classifier part.

\subsection{Data Quantization}

All data and signals were quantized on $n=18$ bits fixed-point representation, which enables to use only 1 dedicated multiplier block (Xilinx DSP Block) for each multiplication. In addition, we perform local quantization by setting the number of integer bits $m \leq n$ at each step of the algorithm. In the subsequent figures depicting hardware blocks, we include the width of each bus in italics. 
The number $m$ of integer bits at each step of the implementation changes as follows:
\begin{itemize}
    \item Feature-vector, Anchor-vector: $m=5$
    \item Distance: $m=10$
    \item Address, Counter: $m=18$
    \item Distance$*$Counter: $m=16$
    \item Anchor-vector$*$Counter: $m=10$
    \item Anchor-vector$+$Feature-vector: $m=10$
\end{itemize}


\subsection{Hardware architecture}
An overview of the hardware architecture is presented in Figure~\ref{fig:HDfig}. Each input feature vector $\textbf{X}^m$ is split into $P$ subvectors, and processed on $P$ Processing blocks in parallel. Each processing block $p$ gets a subvector, as well as an address that is generated by the counter L-P block. Each processing block outputs the class associated to a subvector. The obtained classes $\left(\textbf{c}^p\right)_{1\leq p \leq P}$ are used to compute a Parallelized Majority vote, and classify the input feature vector $\textbf{X}^m$. Finally, Sequential Majority vote is used to output the class of the original signal when data augmentation is performed to classify unlabelled data.


\subsubsection{Processing block}
We use this component to learn or classify a subvector. This component has three inputs: feature subvector, learning-processing signal (L-P), and address (generated by Counter/L-P) and has only one output, the obtained class of a feature subvector one-hot encoded on $C$ bits, where $C$ is the number of classes. Given a feature subvector $\textbf{x}^m_p$, we first compute the euclidean distance between $\textbf{x}^m_p$ and $\textbf{y}^i_p$ (where $\textbf{y}^i$ is the first anchor vector addressed by address generator), multiply the distance by anchor subvector's counter, and store the result in the register $r_p$ in Compare Distance block. We repeat the same process using each $\left(\textbf{y}_p^j\right)_{i\leq j \leq i+k}$, compare the result with the $r_p$ value, and store the smallest one in $r_p$. Finally, Compare Distance block outputs the index of the nearest $\textbf{y}_p^j$ from $\textbf{x}^m_p$. Given this index, Distance register outputs the same index and the class of anchor subvector corresponding to the index. It also outputs a validation signal \textbf{val}, which is equal to $1$ when the nearest $\textbf{y}_p^j$ from $\textbf{x}^m_p$ has been determined. During learning process (L-P=$1$), when \textbf{val} signal is equal to $1$, R-W becomes $0$ and we use the feature subvector and index from Distance Register block through multiplexer to modify the memory content according to Algorithm~\ref{Algo:one}. The inverse of indexes are stored in Look-up tables and multiplied by the output of the Distance Register block (cf. Figure~\ref{fig: Subspace}).

\subsubsection{Counter/L-P}
This component is an ordinary counter, which counts from $0$ to Modulo\_in value. 
Counter/L-P uses a signal (L-P) which is equal to $1$ when learning, and $0$ when test. Modulo\_in value is set to $k$ when learning, to generate only $k$ different addresses in order to read only anchor vectors of a specific class. During test, it is set to $Ck$, in order to read all anchor vectors. 

\subsubsection{Memory}
The Memory block contains two memory blocks (Xilinx UltraRam technology), one to store anchor vectors (URAM A-V), and the other one to store corresponding counters (URAM Counters). Addresses are provided by Counter/L-P. It is also performs the multiplication/division of an anchor vector and its corresponding counter, and the sum between an anchor vector and an input feature vector.
\subsubsection{Majority vote}
Class vectors $\textbf{c}^p$ are one-hot encoded on $C$ bits. Parallel Majority vote computes a bitwise addition over all $\left(\textbf{c}^p\right)_{1\leq p \leq P}$ vectors. The $C$ results are compared sequentially, to obtain the class index $c$ attributed to the unlabelled feature vector $\textbf{x}^m$. 
Sequential Majority vote is computed only when using data augmentation. This block takes as input only one class vector $\textbf{c}^p$ and performs an addition between each $c$ bit of the input class vector and the $c$ inner register. A final comparison is performed between each $c$ results, which outputs a global predicted class vector.

During training, when compare distance block compares two distances, compute distance block computes a new distance between input feature vector and another anchor vector. Thus, the learning phase is $k+3$ clock cycles per feature vector. Precisely, it takes $k$  cycles to compute/compare distances, $1$ cycle to multiply anchor vector with its corresponding counter, $1$ cycle to add the result with the input feature sub vector and increment its counter and $1$ cycle to divide the result by this incremented counter. During classification process, Sequential majority vote needs at least $R$ clock cycles $R$ represents the number of feature vectors resulting form data augmentation) to give an output, Parallel majority vote needs at least $CR$ clock cycles to classify $R$ feature vectors, and Processing block is classifying subvectors of $R$ input feature vector during $CkR$ clock cycles. In the proposed architecture, These three blocks work at the same time, thus $CkR$ cycles are needed to classify an unlabelled feature vector, with $Ck$ cycles to compute distances, repeated $R$ times to classify all feature vectors resulting from data augmentation.   

\begin{table}[h]
    \caption{Accuracy comparaison of TILDA, MLP and SVM}
    \centering
    {\renewcommand{\arraystretch}{1.3}%
    \begin{tabular} { | l | l | l | l | l |} 
 \cline{2-5}
   \multicolumn{1}{l|}{} & TILDA & TILDA-DA & MLP & SVM\\
   
   \hline
  
   CIFAR100 & $69.6\%$ & $65.16\%$ & $68.6\%$ & $67.6\%$ \\
   \hline
   CIFAR10  & $88.7\%$ & $86.6\%$ & $90\%$ & $89.2\%$\\
   \hline
    Imagenet50 & $76\%$ & $74.4\%$ & $75.2\%$ & $75\%$\\
   \hline

    \end{tabular}} \quad
    \vspace{-.3cm}
  
  \label{table:acc}
  \end{table}

\begin{table}[h]
    \caption{FPGA results for the our proposed architecture on vu13p (xcvu13p-figd2104-1-e) ($T=2048$, $P=16$, $K=30$).}
    \centering
    {\renewcommand{\arraystretch}{1.3}%
    \begin{tabular} { | l | l | l |} 
 \cline{2-3}
   \multicolumn{1}{l|}{} & Proposed method & ~\cite{hacene2017incremental}\\
   
   \hline
  
   Memory usage (bits) & $11059488$ & $6553600$ \\
   \hline
   Look-up Tables (LUT) & $152546$ & $95654$\\
   \hline
    DSP & $2064$ & $2048$\\
   \hline
  Maximum frequency (MHz)&$208$ & $204$\\
   \hline
    Learning delay (ns) & $158.2$ & $5$ \\
   \hline
   Classifying delay (ns) & $1442$ & $1470$\\
   \hline
   Energy consumption (W) & $7$ & $13$\\
   \hline
   Accuracy ($\%$) & $87$ & $82$\\
   \hline
  
    \end{tabular}} \quad
    \vspace{-.3cm}
  
  \label{table:Results}
  \end{table}
\section{Results}
\label{sec5}
TILDA achieves an accuracy on par with a MultiLayer Perceptron (MLP) with one hidden layer of 1024 units or a Support Vector Machine classifier (SVM) (c.f. Table~\ref{table:acc}). TILDA-DA is TILDA method without data augmentation and Imagenet50 represents $50$ ImageNet classes which have not been used to train the CNN. Unlike MLP and SVM, TILDA is an incremental method which learns one example at the time, and does not perform an expensive computational backpropagation during training. Moreover, it should be noted that TILDA outperforms other incremental learning methods while using less computational resources~\cite{hacene2018transfer}.

The proposed hardware architecture has been implemented and validated by software simulation over a batch of examples. We provided synthesis results of the architecture on a Xilinx Ultra Scale Vu13p (xcvu13p-figd2104-1-e) Field Programmable Gate Array (FPGA) (Table \ref{table:Results}). 

Performance estimates are given for the CIFAR10 for $P=16$, $K=30$ and $C=10$, yielding an accuracy of $89\%$/$87\%$ with/without data augmentation, instead of $88.7\%$/$86.6\%$ obtained for $32$ bits encoding. To obtain feature vectors, we use inception V3~\cite{szegedy2015rethinking} ($T=2048$). $2048$ DSPs are used to compute distances and $P=16$ more to multiply/divide anchor vectors by their corresponding counters. Power consumption and maximum clock frequency of the whole system are estimated to about $8$ Watts and $208$ MHz. The estimated time needed to learn/classify an input vector is $158.2$/$1442$ ns at maximum clock frequency, corresponding to an acceleration factor of $10^4$ when compared with a software simulation delay using an I7 870 (2.93 GHz) processor.


\section{Conclusion}
\label{sec6}
In this paper we introduced an architecture for incremental learning on chip. As such, the proposed method is able to learn one example at a time, and does not require to store the whole dataset to perform training.
The proposed architecture allows an embedded system to accommodate new data from scratch and classify previously unlabelled inputs, and more important, performs the learning on chip. Future work will introduce hardware architecture and implementation of the pretrained CNN to propose a complete embedded solution.

\bibliographystyle{IEEEtran}
\bibliography{IEEEabrv,reb}

\end{document}